%% file: Paper.tex
\documentclass[a4paper,twoside]{article}

\usepackage{epsfig}
\usepackage{subcaption}
\usepackage{calc}
\usepackage{amssymb}
\usepackage{amstext}
\usepackage{amsmath}
\usepackage{amsthm}
\usepackage{multicol}
\usepackage{pslatex}
\usepackage{apalike}
\usepackage{algorithm2e}

\usepackage[bottom]{footmisc}

\usepackage[dvipsnames]{xcolor}
\usepackage{wrapfig}
\usepackage{microtype}
\usepackage{graphicx}
\usepackage{subcaption}
\usepackage{amsmath}
\usepackage{amssymb}
\usepackage{mathtools}
\usepackage{amsthm}
\usepackage{multirow}
\usepackage{makecell}
\usepackage{threeparttable}
\usepackage{pifont}
\usepackage{algorithmic}
\usepackage{algorithm}
\usepackage{amsmath}
\usepackage{amssymb}
\usepackage{booktabs}
\usepackage{apalike}
\let\bibhang\relax
\usepackage{natbib}
\DeclareMathAlphabet{\mathcal}{OMS}{cmsy}{m}{n}

\newcommand{\methodname}{HASTE}
\newcommand{\cmark}{\ding{51}}
\newcommand{\xmark}{\ding{55}}
\input{math_commands.tex}

\usepackage{algorithmic}
\usepackage{algorithm}
\usepackage{url}

\usepackage{SCITEPRESS}     % Please add other packages that you may need BEFORE the SCITEPRESS.sty package.

\begin{document}

\title{Data-Free Dynamic Compression of CNNs for Tractable Efficiency}

\author{\authorname{Lukas Meiner\sup{1,2}\orcidAuthor{0009-0003-1451-2197}, Jens Mehnert\sup{1}\orcidAuthor{0000-0002-0079-0036} and Alexandru Paul Condurache\sup{1,2}\orcidAuthor{0000-0002-0626-335X}}
	\affiliation{\sup{1}Cross-Domain Computing Solutions, Robert Bosch GmbH, Daimlerstraße 6, 71229 Leonberg, Germany}
	\affiliation{\sup{2}Institute for Signal Processing, Universität zu Lübeck, Ratzeburger Allee 160, 23562 Lübeck, Germany}
	\email{\{Lukas.Meiner, JensEricMarkus.Mehnert, AlexandruPaul.Condurache\}@bosch.com}
}


\input{0_abstract.tex}

\onecolumn \maketitle \normalsize \setcounter{footnote}{0} \vfill

\input{1_introduction.tex}
\input{2_relatedwork.tex}
\input{3_method.tex}
\input{4_experiments.tex}
\input{5_conclusion.tex}

\bibliographystyle{apalike}
{\small
\bibliography{references.bib}}

\appendix
\section*{\uppercase{Appendix}}
\input{6_appendix.tex}

\end{document}

%% file: math_commands.tex
\usepackage{amsmath,amsfonts,bm}

\def\eqref#1{equation~\ref{#1}}
\def\1{\bm{1}}

\DeclareMathAlphabet{\mathsfit}{\encodingdefault}{\sfdefault}{m}{sl}
\SetMathAlphabet{\mathsfit}{bold}{\encodingdefault}{\sfdefault}{bx}{n}

\def\sN{{\mathbb{N}}}

\def\sR{{\mathbb{R}}}

\newcommand{\normlone}{L^1}

%% file: 0_abstract.tex
\abstract{
To reduce the computational cost of convolutional neural networks (CNNs) on resource-constrained devices, structured pruning approaches have shown promise in lowering floating-point operations (FLOPs) without substantial drops in accuracy. 
However, most methods require fine-tuning or specific training procedures to achieve a reasonable trade-off between retained accuracy and reduction in FLOPs, adding computational overhead and requiring training data to be available.
To this end, we propose \methodname\ (\textbf{Has}hing for \textbf{T}ractable \textbf{E}fficiency), a data-free, plug-and-play convolution module that instantly reduces a network's test-time inference cost without training or fine-tuning.
Our approach utilizes locality-sensitive hashing (LSH) to detect redundancies in the channel dimension of latent feature maps, compressing similar channels to reduce input and filter depth simultaneously, resulting in cheaper convolutions.
We demonstrate our approach on the popular vision benchmarks CIFAR-10 and ImageNet, where we achieve a 46.72\% reduction in FLOPs with only a 1.25\% loss in accuracy by swapping the convolution modules in a ResNet34 on CIFAR-10 for our \methodname\ module.
}

%% file: 1_introduction.tex
\section{\uppercase{Introduction}}

\begin{figure*}[h!]
	\begin{center}
		\includegraphics[width=0.9\linewidth]{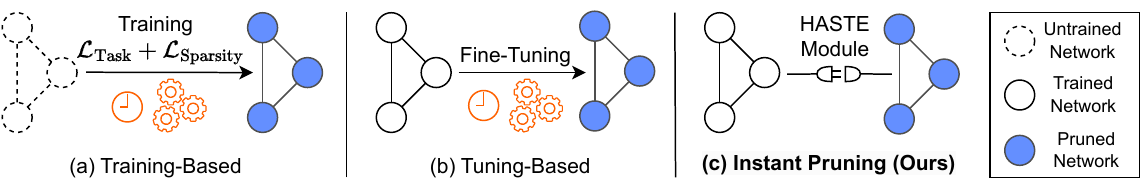}
	\end{center}
	\caption{Overview of related pruning approaches. Training-based methods require specialized training procedures. Methods based on fine-tuning need retraining to compensate lost accuracy in the pruning step. Our method instantly reduces network FLOPs and maintains high accuracy entirely without training or fine-tuning.}
	\label{figure_overview}
\end{figure*}

With the rise in availability and capability of deep learning hardware, the possibility to train ever larger models led to impressive achievements in the field of computer vision. At the same time, concerns regarding high computational costs, environmental impact and the applicability on resource-constrained devices are growing. This led to the introduction of carefully constructed efficient models \citep{Howard2017MobileNets, Sandler2018MobileNetV2, Tan2019EfficientNet, Tan2021EfficientNetV2, Zhang2018ShuffleNet, Ma2018ShuffleNet} that offer fast inference in embedded applications, gaining speed by introducing larger inductive biases. Yet, highly scalable and straight-forward architectures \citep{Simonyan2015Very, He2016Deep, Dosovitskiy2021Image, Liu2021Swin, Liu2022ConvNet, Woo2023ConvNeXt} remain popular due to their performance and ability to generalize, despite requiring more data, time and energy to train. To still allow for larger models to be used in mobile applications, various methods \citep{Zhang2016Accelerating, Lin2017Towards, Pleiss2017Memory, Han2020GhostNet, Luo2017ThiNet} have been proposed to reduce their computational cost. One particularly promising field of research for the compression of convolutional architectures is pruning \citep{Wimmer2023Dimensionality}, especially in the form of structured pruning for direct resource savings \citep{Anwar2017Structured}.

However, the application of existing work is restricted by two factors. Firstly, many proposed approaches rely on actively learning which channels to prune during the regular model training procedure \citep{Dong2017More, Liu2017Learning, Gao2019Dynamic, Verelst2020Dynamic, Bejnordi2020Batch, Li2021Dynamic, Xu2021Efficient}. This introduces additional parameters to the model, increases the complexity of the optimization process due to supplementary loss terms, and requires existing models to be retrained to achieve any reduction in FLOPs. The second limiting factor is the necessity of performing fine-tuning steps to restore the performance of pruned models back to acceptable levels \citep{Wen2016Learning, Li2017Pruning, Lin2017Runtime, Zhuang2018Discrimination, He2018AMC}. Aside from the incurred additional cost and time requirements, this creates a dependency on the availability of the data that was originally used to train the baseline model, as tuning the model on a different set of data can lead to catastrophic forgetting \citep{Goodfellow2014Empirical}.

To this end, we propose \methodname, a plug-and-play channel pruning approach that is entirely data-free and does not require any real or synthetic training data. Our method instantly reduces the computational complexity of convolution modules without requiring any additional training or fine-tuning. To achieve this, we utilize a locality-sensitive hashing scheme \citep{Indyk1998Approximate} to dynamically detect and cluster similarities in the channel dimension of latent feature maps in CNNs. By exploiting the distributive property of the convolution operation, we take the average of all input channels that are found to be approximately similar and convolve it with the sum of corresponding filter channels. This reduced convolution is performed on a smaller channel dimension, which drastically lowers the amount of FLOPs required. The trade-off between retained accuracy and compression ratio is directly steerable by altering one hyperparameter shared across all \methodname\ modules in the network, which simplifies experimentation for users.

Our experiments demonstrate that the \methodname\ module is capable of greatly reducing computational cost of a wide variety of pre-trained CNNs while maintaining high accuracy. More importantly, it does so directly after exchanging the original convolutional modules for the \methodname\ block. This allows us to skip lengthy model trainings with additional regularization and sparsity losses as well as extensive fine-tuning procedures. Furthermore, we are not tied to the availability of the dataset on which the given model was originally trained. Our pruning approach is entirely data-free, thus enabling pruning in a setup where access to the trained model is possible, but access to the data is restricted. Finally, this allows us to adjust the computational cost of a model in real time, adapting its test-time complexity to the availability of hardware resources. To the best of our knowledge, this makes the \methodname\ module the first dynamic and data-free CNN pruning approach that does not require any form of training or fine-tuning.

Our main contributions are:
\begin{itemize}
	\item We propose a locality-sensitive hashing based method to dynamically detect redundancies in the latent features of current CNN architectures. Our method incurs a low computational overhead and is entirely data-free.
	\item We propose \methodname, a scalable, plug-and-play convolution module replacement that leverages these structural redundancies to save computational complexity in the form of FLOPs at test time, without requiring any training steps.
	\item We showcase our method's performance on popular CNN models trained on benchmark vision datasets. We also identify a positive scaling behavior, achieving higher cost reductions on deeper and wider models.
\end{itemize}

%% file: 2_relatedwork.tex
\section{\uppercase{Related Work}}
When structurally pruning a model, its computational complexity is reduced at the expense of performance on a given task. For this reason, fine-tuning is often performed after the pruning scheme was applied. The model is trained again in its pruned state to compensate the loss of structural components, often requiring multiple epochs of tuning \citep{Li2017Pruning, Zhuang2018Discrimination, Xu2021Efficient} on the training dataset. These methods tend to remove structures from models in a static way, not adjusting for different degrees of sparsity across varying input data. Some recent methods avoid fine-tuning by learning a pruning pattern during regular model training \citep{Liu2017Learning, Gao2019Dynamic, Xu2021Efficient, Li2021Dynamic, Elkerdawy2022Fire}. This generates an input-dependent dynamic path through the network, allocating less compute to sparser images. 

\textbf{Static Pruning.}
By finding general criteria for the importance of individual channels, some recent methods propose static pruning approaches. PFEC \citep{Li2017Pruning} prunes filter kernels with low importance measured by their $\normlone$-norm in a one-shot manner. DCP \citep{Zhuang2018Discrimination} equips models with multiple loss terms before fine-tuning to promote highly discriminative channels to be formed. Then, a channel selection algorithm picks the most informative ones. FPGM \citep{He2019Filter} demonstrates a fine-tuning-free pruning scheme, exploiting norm-based redundancies to train models with reduced complexity. AMC \citep{He2018AMC} explores a compression policy generated by reinforcement learning. A handful of data-free approaches exist, yet they either use synthetic data to retrain the model \citep{Yin2020Dreaming} or generate a static model \citep{Yvinec2023RED, Bai2023Unified} that is unable to adapt its compression to the availability of hardware resources on the fly. We target the dynamic compression of models in a data-free manner.

\textbf{Dynamic Gating.}
To accommodate inputs of varying complexity in the pruning process, recent works try to learn dynamic, input-dependent paths through the network \citep{Xu2021Efficient, Li2021Dynamic, Elkerdawy2022Fire, Liu2017Learning, Hua2019Channel, Verelst2020Dynamic, Bejnordi2020Batch, Liu2019Dynamic}. These methods learn (binary) masks that toggle structural components of the underlying CNN at runtime. This requires storing all of the model's weights, as each weight is potentially important for specific inputs. 
DGNet \citep{Li2021Dynamic} equips the base model with additional spatial and channel gating modules based on average pooling that are trained end-to-end together with the model using additional regularization losses. Similarly, DMCP \citep{Xu2021Efficient} learns mask vectors using a pruning loss and does not need fine-tuning procedures after training.
FTWT \citep{Elkerdawy2022Fire} decouples the task and regularization losses introduced by previous approaches, reducing the complexity of the pruning scheme.
While these methods do not require fine-tuning, they introduce additional complexity through pruning losses and the need for custom gating modules during training to realize FLOP savings. We focus on real-time compression during model inference, with no training and data requirement at all. This also enables us to have tractable compression ratios at test time, as we do not require training towards a set ratio.

\textbf{Hashing for Efficient Inference.} In recent years, the usage of locality-sensitive hashing \citep{Indyk1998Approximate} schemes as a means to make model inference more efficient has gained some popularity. Reformer \citep{Kitaev2020Reformer} uses LSH to reduce the computational complexity of multi-head attention modules in transformer models by finding similar queries and keys before computing their matrix product. \citet{Mueller2022} employ a multiresolution hash encoding to construct an efficient feature embedding for neural radiance fields (NeRFs), leading to orders of magnitude speedup compared to previous methods. 
SLIDE \citep{Chen2020SLIDE} and MONGOOSE \citep{Chen2021MONGOOSE} use a similar LSH scheme to store non-contiguous activation patterns of a high-dimensional feedforward network, only computing the strongest activating neurons during the forward pass. Using specialized C++ and CUDA code, the authors achieve significant speedups on CPUs as well as GPUs. Other approaches related to LSH have also been explored for model compression. \citet{Liu2021Efficient} employ a count sketch-type algorithm to approximate the forward pass of multilayer perceptrons by hashing the model's input vector. FPKM \citep{Liu2021More} extends on FPGM \citep{He2019Filter} and explores the use of $k$-means clustering for finding redundant input channels. However, this approach is limited to fixed pruning ratios determined by the amount of clusters, and does not allow for dynamic compression.

%% file: 3_method.tex
\section{\uppercase{Method}}
In this section, we present \methodname, a novel convolution module based on locality-sensitive hashing that acts as a plug-and-play replacement for any regular convolution module, instantly reducing the FLOPs during inference. 
Firstly, we give a formal definition of the underlying LSH scheme.
Secondly, we illustrate how hashing is used to identify redundancies inside latent features of convolutional network architectures. 
Lastly, we present the integration of the hashing process into our proposed \methodname\ module, which allows us to compress latent features for cheaper computations. 

\subsection{Locality-Sensitive Hashing via Sparse Random Projections}
\label{method_finding_red}
\begin{figure*}[th!]
	\centering
	\includegraphics[width=1.0\linewidth]{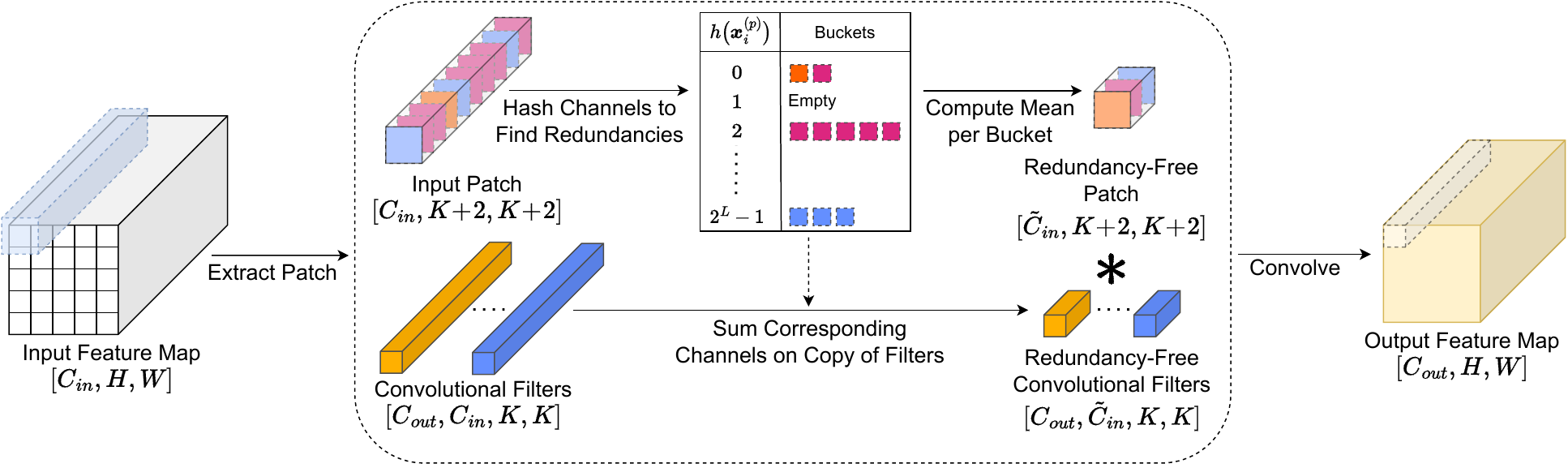}
	\caption{Overview of our proposed \methodname\ module. Each patch of the input feature map is processed to find redundant channels. Detected redundancies are then merged together, dynamically reducing the depth of each patch and the convolutional filters.}
	\label{figure_method}
\end{figure*}

\label{method_LSH}
Locality-sensitive hashing is a popular approach for approximate fast nearest neighbor search in high-dimensional spaces. A hash function $h: \sR^d \to \sN$ is locality-sensitive, if similar vectors in the input domain $x, y \in \sR^d$ receive the same hash codes $h(x) = h(y)$ with high probability. This is in contrast to regular hashing schemes which try to reduce hash collisions to a minimum, widely scattering the input data across their hash buckets. 
More formally, we require a measure of similarity on the input space and an adequate hash function $h$. A particularly suitable measure for use in convolutional architectures is the cosine similarity, as convolving the (approximately) normalized kernel with the normalized input is equivalent to computing their cosine similarity. Pairwise similarities between vectors are preserved through hashing by the allocation of similar hash codes. 

One particular family of hash functions that groups input data by cosine similarity is given by random projections (RP). These functions partition the high-dimensional input space through $L$ random hyperplanes, such that each input vector is assigned to exactly one section of this partitioning, called a hash bucket. Determining the position of an input $x \in \sR^d$ relative to all $L$ hyperplanes is done by computing the dot product with their normal vectors $v_l \in \sR^d,\, l \in \{1, \dots, L\}$, whose entries are drawn from a standard normal distribution $\mathcal{N}(0,1)$. 
By defining
\begin{equation}
	h_l: \sR^d \to \{0,1\},\,\,\, h_l(x) := \begin{cases}
		1,\, \text{if } v_l \cdot x > 0,\\
		0,\, \text{else,}
	\end{cases}
\end{equation}
we get a binary information representing to which side of the $l$-th hyperplane input $x$ lies. The hyperparameter $L$ governs the discriminative power of this method, dividing the input space $\sR^d$ into a total of $2^L$ distinct regions, or hash buckets. By concatenating all individual functions $h_l$, we receive the RP hash function 
\begin{equation}
	h:\sR^d \to \{0,1\}^L,\,\,\, h(x) = \left(h_1(x), \dots, h_L(x)\right).
\end{equation}
Note that $h(x)$ is an $L$-bit binary code, acting as an identifier of exactly one of the $2^L$ hash buckets. Equivalently, we can transform this code into an integer, labeling the hash buckets from $0$ to $2^{L}-1$:
\begin{equation}
\begin{gathered}
		h:\sR^d \to \left\{0, \dots, 2^L-1\right\} \\
		h(x) = 2^{L-1}h_L(x) + \dots + 2^0h_1(x).
\end{gathered}
\end{equation}

While LSH already reduces computational complexity drastically compared to exact nearest neighbor search, the binary code generation still requires $L \cdot d$ multiplications and $L \cdot (d-1)$ additions per input. To further decrease the cost of this operation, we employ the method presented by \citep{Achlioptas2003Database, Li2006Very}: Instead of using standard normally distributed vectors $v_l$, we use very sparse vectors $\tilde{v}_l$, containing only elements from the set $\{1,0,-1\}$. Given a targeted degree of sparsity $s \in (0,1)$, the hyperplane normal vectors $\tilde{v}_l$ are constructed randomly such that the expected ratio of zero entries is $s$. The remaining $1-s$ of vector components are randomly filled with either $1$ or $-1$, both chosen with equal probability. 
This reduces the dot product computation to a total of $L\cdot (d(1-s)-1)$ additions and $0$ multiplications, as we only need to sum entries of $x$ where $\tilde{v}_l$ is non-zero with the corresponding signs. Consequently, this allows us to trade expensive multiplication operations for cheap additions.

\subsection{Finding Redundancies with LSH}
After establishing LSH via sparse random projections as a computationally cheap way to find approximate nearest neighbors in high-dimensional spaces, we now aim to leverage this method as a means of finding redundancies in the channel dimension of latent feature maps in CNNs.  
Formally, a convolutional layer can be described by sliding multiple learned filters $F_j \in \sR^{C_{in} \times K \times K},\, j \in \{1, \dots, C_{out}\}$ over the (padded) input feature map $X \in \sR^{C_{in} \times H \times W}$ and computing the discrete convolution at every point. Here, $K$ is the kernel size, $H$ and $W$ denote the spatial dimensions of the input, and $C_{in}, C_{out}$ describe the input and output channel dimensions, respectively. 

For any filter position, the corresponding input window contains redundant information in the form of similar channels. However, a regular convolution module ignores potential savings from reducing the amount of similar computations in the convolution process.
We challenge this design choice and instead leverage redundant channels to save computations in the convolution operation. As the first step, we rasterize the (padded) input feature map into patches $X_i^{(p)} \in \sR^{(K+2) \times (K+2)}$ for $i = 1, \dots, C_{in}$, with an overlap of two pixels on each side. This is equivalent to splitting the spatial dimension into patches of size $K \times K$, but keeping the filter overlap to its neighbors. The special case of $K=1$ is discussed in Appendix \ref{appendix_1x1_conv}. 

To group similar channels together, we flatten all individual channels $X^{(p)}_i$ into vectors of dimension $(K+2)^2$ and center them by the mean along the channel dimension for any given patch $p$. We denote the resulting vectors as $x_i^{(p)}$. Finally, they are hashed using $h$, giving us a total of $C_{in}$ hash codes.
We then check which hash code appears more than once, as all elements that appear in the same hash bucket are determined to be approximately similar by the LSH scheme. Consequently, grouping the vector representations of $X^{(p)}_i$ by their hash code, we receive sets of redundant feature map channels.

In particular, note that our RP LSH approach is invariant to the scaling of a given input vector. This means that input channels of the same spatial structure, but with different activation intensities, still land in the same hash bucket, effectively finding even more redundancies in the channel dimension.

\subsection{The \methodname\ Module}
\label{method_our_module}
Our approach is motivated by the distributivity of the convolution operation. Instead of convolving various filter kernels with nearly similar input channels and summing the result, we can approximate this operation by computing the sum of kernels first and convolving it with the mean of these redundant channels. 
The grouping of input channels $X^{(p)}_i$ into hash buckets provides a straight-forward way to utilize this distributive property for the reduction of required floating-point operations when performing convolutions.

To avoid repeated computations on nearly similar channels, we dynamically reduce the size of each input context window  $X^{(p)}$ by compressing channels found in the same hash bucket, as shown in Figure \ref{figure_method}. The merging operation is performed by taking the mean of all channels in one bucket. As a result, the number of remaining input channels of a given patch is reduced to $\tilde{C}_{in} < C_{in}$.
In a similar manner to the reduction of the input feature map depth, we add the corresponding channels of all convolutional filters $F_j$. Note that this does not require hashing of the filter channels, as we can simply aggregate those kernels that correspond to the collapsed input channels. 
This step is done on the fly for every patch $p$, retaining the original filter weights for the next patch. 

The choice of different merging operations for input and filter channels is directly attributable to the distributive property, as the convolution between the average input and summed filter set retains a similar output intensity to the original convolution. When choosing to either average or sum both inputs and filters, we would systematically under- or overestimate the original output, respectively.

Finally, the reduced input patch is convolved with the reduced set of filters in a sliding window manner to compute the output. This can be formalized as follows:
\begin{equation}
	\begin{gathered}
	\label{eq_approximation_of_convolution}
		\sum_{i=1}^{C_{in}} F_{j,i} * X_i^{(p)} \approx \\
		\sum_{\substack{l=0 \\ S_l^{(p)} \neq \emptyset}}^{2^L-1} 
		\Biggl( 
		\biggl( \,\, \sum_{i \in S_l^{(p)}} F_{j,i}\biggr) * 
		\biggl( \frac{1}{|S_l^{(p)}|} \sum_{i \in S_l^{(p)}} X_i^{(p)}\biggr) 
		\Biggr) \,, 
	\end{gathered}
\end{equation}
where $S_l^{(p)} = \{i \in \{1, \dots, C_{in}\} \, \vert \, h(x_i^{(p)}) = l\}$ contains all channel indices that appear in the $l$-th hash bucket. 
%We assume the scaling factor $1/|S_l^{(p)}|$ to be zero if the bucket is empty for simplicity of notation. 
Since we do not remove entire filters, but rather reduce their depth, the output feature map retains the same spatial dimension and number of channels as with a regular convolution module. The entire procedure is summarized in Algorithm \ref{method_algo}.

This reduction of input and filter depth lets us define a compression ratio $r = 1-(\tilde{C}_{in} / C_{in}) \in (0, 1)$,
determining the relative reduction in channel depth. 
Note that this ratio is dependent on the amount of redundancies in the input feature map $X$ at patch position $p$. Our dynamic pruning of channels allows for different compression ratios across images and even in different regions of the same input.

Although the hashing and merging operations create additional computational cost, the overall savings on computing the convolution operations with reduced channel dimension outweigh the added overhead. The main additional cost lies in the merging of filter channels, as this process is repeated $C_{out}$ times for every patch $p$. However, since this step is performed by computationally cheap additions, it lends itself to hardware-friendly implementations.

\begin{table}
	\centering
	\caption{Overview of related pruning approaches. While other methods require either fine-tuning or a specialized training procedure to achieve notable FLOPs reduction, our method is completely training-free and data-free.}
	\label{tab_other_approaches_overview}
	\resizebox{1.0\columnwidth}{!}{%
		\begin{tabular}{lcccc}
			\toprule
			\multirow[c]{2.5}{*}{Method} & \multirow[c]{2.5}{*}{Dynamic} & \multicolumn{3}{c}{Restrictive Requirements} \\
			\cmidrule{3-5}
			& & Training & Fine-Tuning & Data Availability   \\
			\midrule
			SSL \citep{Wen2016Learning} & \xmark & \xmark & \cmark & \cmark \\
			PFEC \citep{Li2017Pruning} & \xmark & \xmark & \cmark & \cmark \\
			LCCN \citep{Dong2017More} & \cmark & \cmark & \xmark & \cmark \\
			FBS \citep{Gao2019Dynamic} & \cmark & \cmark & \xmark & \cmark \\
			FPGM \citep{He2019Filter} & \xmark & \cmark & \xmark & \cmark \\
			DynConv \citep{Verelst2020Dynamic} & \cmark & \cmark & \xmark & \cmark \\
			DMCP \citep{Xu2021Efficient} & \cmark & \cmark & \xmark & \cmark \\
			DGNet \citep{Li2021Dynamic} & \cmark & \cmark & \xmark & \cmark \\
			FTWT \citep{Elkerdawy2022Fire} & \cmark & \cmark & \xmark & \cmark \\
			\midrule
			\textbf{\methodname\ (ours)} & \cmark & \xmark & \xmark & \xmark \\
			\bottomrule 
		\end{tabular}%
	}
\end{table}

\begin{algorithm}%[H]
	\footnotesize
	\caption{Pseudocode overview of the \methodname\ module.}
	\label{method_algo}
	\begin{algorithmic}
		\STATE \textbf{Input}: Feature map $X \in \sR^{C_{in} \times H \times W}$,\\ 
		Filters $F \in \sR^{C_{out} \times C_{in} \times K \times K}$ \\
		\textbf{Output}: $Y \in \sR^{C_{out} \times H \times W}$ \\
		\textbf{Initialize:} $h: \sR^{(K+2)^2} \to \{0, \dots, 2^L-1\}$ 
		\FOR{every patch $p$}
		\STATE HashCodes = [ ]
		\FOR{$i = 1, \dots, C_{in}$}
		\STATE $x_i^{(p)} = $ Center(Flatten($X^{(p)}_i$))
		\STATE HashCodes.Append($h(x_i^{(p)})$)
		\ENDFOR
		\STATE $\tilde{X}^{(p)}$ = MergeInput($X^{(p)}$, HashCodes) 
		\STATE $\tilde{F}$ = MergeFilters($F$, HashCodes) 
		\STATE $Y^{(p)}$ = $\tilde{X}^{(p)}$ * $\tilde{F}$
		\ENDFOR
		\STATE \textbf{return} $Y$
	\end{algorithmic}
\end{algorithm}

Our \methodname\ module features two hyperparameters: the number of hyperplanes $L$ in the LSH scheme and the degree of sparsity $s$ in their normal vectors. Adjusting $L$ gives us a tractable trade-off between the compression ratio and retained accuracy. This allows us to generate multiple model variants from one underlying base model, either focusing on low FLOPs or high accuracy. The normal vector sparsity $s$ does not require direct tuning and can easily be fixed across a dataset. \citet{Achlioptas2003Database} and \cite{Li2006Very} provide initial values with theoretical guarantees. Our hyperparameter choices are discussed in Section \ref{sec_experiment_settings}.

%% file: 4_experiments.tex
\section{\uppercase{Experiments}}
\label{sec_experiments}
In this section, we present results of our plug-and-play approach on standard CNN architectures in terms of FLOPs reduction as well as retained accuracy. Firstly, we describe the setup of our experiments in detail. Then, we evaluate our proposed \methodname\ module on the CIFAR-10 \citep{Krizhevsky2009Learning} dataset for image classification and discuss the influence of the hyperparameter $L$. Lastly, we present results on the ImageNet ILSVRC 2012 \citep{Russakovsky2015ImageNet} benchmark and discuss the scaling behavior of our method.

\begin{figure*}
	\centering
	\includegraphics[width=1\linewidth]{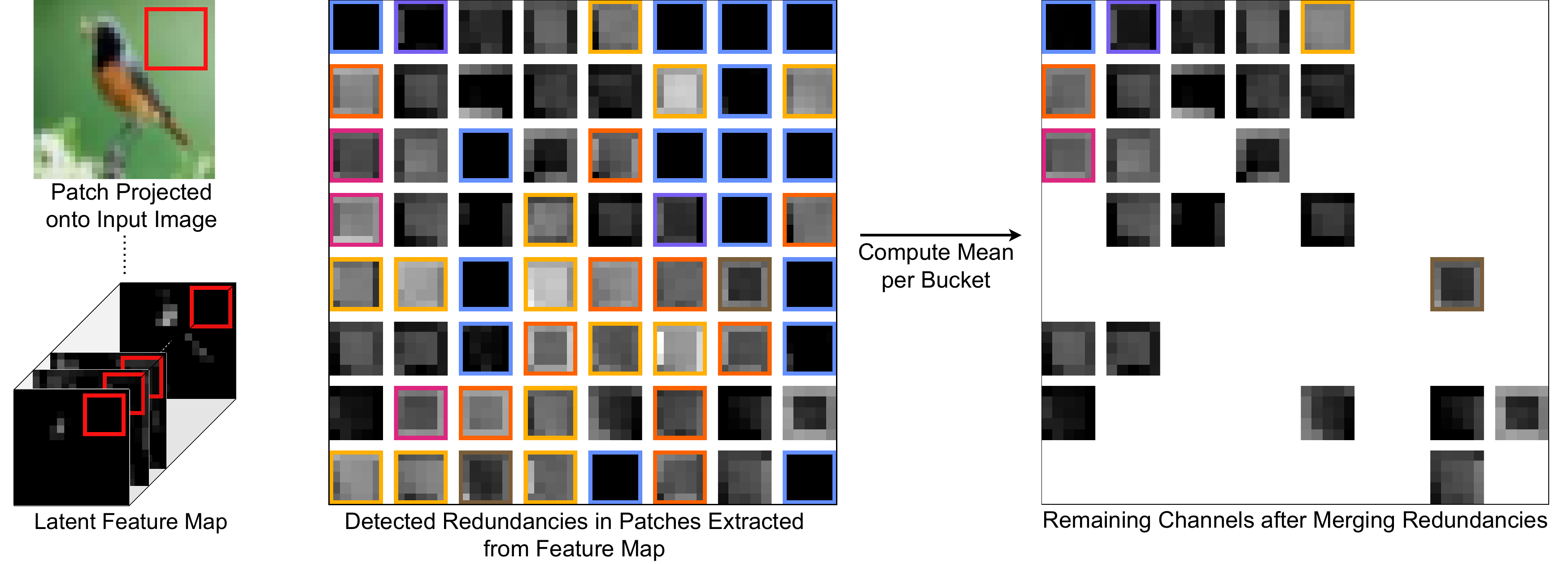}
	\caption{Visualization of the input channel compression performed by the \methodname\ module in a ResNet18 model on CIFAR-10. One observed patch is marked as a red square on the input feature maps. All 64 channels of this patch are then plotted in an $8 \times 8$ grid. Patches with identical hash codes receive identical outline colors and are averaged by taking their mean. Patches with no matching hash code are left unchanged. Here, we reduce the input channel dimension from $64$ to $24$, which gives us a compression ratio of $r = 62.50\%$.}
	\label{fig_fm_vis_1}
\end{figure*}

\subsection{Experiment Settings}
\label{sec_experiment_settings}
For the experiments on CIFAR-10, we used pre-trained models provided by \citep{Phan2021huyvnphan/PyTorch_CIFAR10}. On ImageNet, we use the trained models provided by PyTorch $2.0.0$ \citep{Paszke2019PyTorch}. Given a baseline model, we replace the regular non-strided convolutions with our \methodname\ module. For ResNet models \citep{He2016Deep}, we do not include downsampling layers in our pruning scheme. 

Depending on the dataset, we vary the degree of sparsity $s$ in the hyperplanes as well as at which layer we start pruning. As the CIFAR-10 dataset is less complex and features smaller latent spatial dimensions, we can increase the sparsity and prune earlier compared to models trained on ImageNet. For this reason, we set $s = 2/3$ on CIFAR-10 experiments as suggested by \citet{Achlioptas2003Database}, and start pruning VGG models \citep{Simonyan2015Very} from the first convolution module and ResNet models from the first block after the max pooling operation. For experiments on ImageNet, we choose $s = 1/2$ to create random hyperplanes with less non-zero entries, leading to a more accurate hashing scheme. VGG models are pruned starting from the third convolution module and ResNet / WideResNet models starting from the second layer. These settings compensate the lower degree of redundancy in latent feature maps of ImageNet models, especially in the early layers. A detailed component ablation of our method is found in Appendix \ref{appendix_ablation}.

After plugging in our \methodname\ modules, we directly evaluate the models on the corresponding test set using one NVIDIA Tesla T4 GPU on an internal cluster, as no further fine-tuning or retraining is required. We follow common practice and report results on the validation set of the ILSVRC 2012 for models trained on ImageNet. 
Each experiment is repeated for three different random seeds to evaluate the effect of random hyperplane initialization. We report the mean top-1 accuracy after pruning and the mean FLOPs reduction compared to the baseline model as well as the standard deviation for both values. Additionally, we provide latency estimates for the proposed \methodname\ module in Tables \ref{tab_latency_c10} and \ref{tab_latency_imagenet}, measured on an Intel i7-11850H CPU. For more details on the latency, we refer to the Appendix \ref{appendix_latency}.

Since, to the best of our knowledge, \methodname\ is the only approach that offers entirely data-free and dynamic model compression, we cannot give a direct comparison to similar work. For this reason, we resort to showing results of related channel pruning and dynamic gating approaches that feature specialized training or tuning routines. An overview of these methods is given in Table \ref{tab_other_approaches_overview}.

\subsection{Results on CIFAR-10}
\begin{figure}
	\centering
	\begin{subfigure}{0.45\textwidth}
		\centering
		\includegraphics[width=0.95\textwidth]{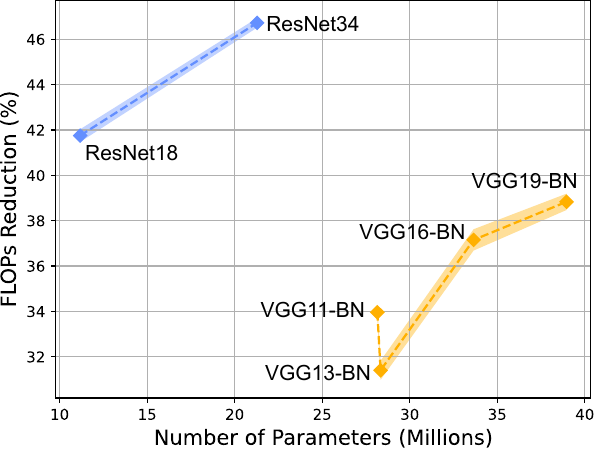}
		\caption{Overview of CIFAR-10 results.}
		\label{subfig_cifar10_overview}
	\end{subfigure}
	\par\bigskip\bigskip
	\begin{subfigure}{0.45\textwidth}
		\centering
		\includegraphics[width=0.95\textwidth]{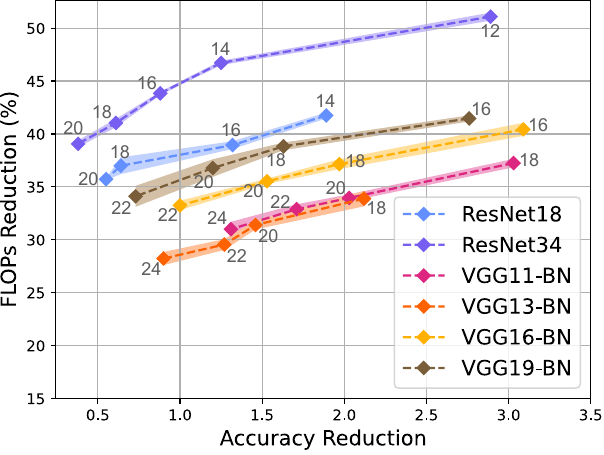}
		\caption{Influence of hyperparameter $L$.}
		\label{subfig_cifar10_pareto}
		\vspace*{0.25cm}
	\end{subfigure}
	\caption{Results of our method on the CIFAR-10 dataset. (a) shows the achieved FLOPs reduction
		for all tested models, using $L = 14$ for ResNets and $L = 20$ for VGG-BN models. (b) depicts the
		influence of the chosen number of hyperplanes $L$ (shown in \textcolor{gray}{gray}) on compression rates and accuracy.}
\end{figure}

For the CIFAR-10 dataset, we evaluate our method on ResNet18 and ResNet34 architectures as well as on VGG11-BN, VGG13-BN, VGG16-BN and VGG19-BN. Results are presented in Figure \ref{subfig_cifar10_overview}. To gain an intuitive understanding of our proposed \methodname\ module, we visualize the LSH-based channel clustering in Figure \ref{fig_fm_vis_1}. Further visualizations are provided in Appendix \ref{appendix_visualizations}.
Overall, our method achieves substantial reductions in the FLOPs requirement of tested networks. In particular, it reduces the computational cost of a ResNet34 by 46.72\% entirely without training, while only losing 1.25 percentage points accuracy.

The desired ratio of cost reduction to accuracy loss can be adjusted on the fly by changing the hyperparameter $L$ across all \methodname\ modules simultaneously. Figure \ref{subfig_cifar10_pareto} shows how the relationship of targeted cost reduction and retained accuracy is influenced by the choice of $L$. Increased accuracy on the test set, achieved by increasing $L$, is directly related to less FLOPs reduction. For instance, we can vary the accuracy loss on ResNet34 between 2.89 ($L=12$) and 0.38 ($L=20$) percentage points to achieve 51.09\% and 39.07\% reduction in FLOPs, respectively.

We also give an overview of results from related approaches in Table \ref{tab_results_c10}. Although our method is not trained or fine-tuned on the dataset, it achieves comparable results to approaches which tailored their pruning scheme to the data. 
Specifically, for the ResNet18 and VGG19-BN models, our method is on par with the best trained approaches, namely DMCP \citep{Xu2021Efficient} and SSL \citep{Wen2016Learning}, achieving a similar ratio of FLOPs reduction to retained accuracy.

\begin{table}[h]
	\centering
	\caption{Selected results on CIFAR-10. "FLOPs Red." denotes the percentage decrease of FLOPs after pruning compared to the base model.}
	\label{tab_results_c10}
	\resizebox{1\linewidth}{!}{%
		\begin{threeparttable}
			\begin{tabular}{ccccccc}
				\toprule
				\multirow[c]{2.5}{*}{Model} & \multirow[c]{2.5}{*}{Method} & \multicolumn{3}{c}{Top-1 Accuracy (\%)} & \multirow[c]{2.5}{*}{\makecell{FLOPs Red.\ \\(\%)}} & \multirow[c]{2.5}{*}{\makecell{Data-\\Free}}\\
				\cmidrule(lr){3-5}
				& & Baseline & Pruned & $\Delta$ & & \\
				\midrule
				\multirow[c]{5.5}{*}{\rotatebox{90}{ResNet18}} & PFEC$^*$ & 91.38 & 89.63 & 1.75 & 11.71 & \xmark \\
				& SSL$^*$ & 92.79 & 92.45 & 0.34 & 14.69 & \xmark \\ 
				& DMCP & 92.87 & 92.61 & 0.26 & 35.27 & \xmark  \\ 
				\cmidrule{2-7}
				& \textbf{Ours} ($L=14$) & 93.07 & 91.18 $(\pm 0.38)$ & 1.89 & 41.75 $(\pm 0.28)$ & \cmark \\
				& \textbf{Ours} ($L=20$) & 93.07 & 92.52 $(\pm 0.10)$ & 0.55 & 35.73 $(\pm 0.09)$ & \cmark \\
				\midrule
				\multirow[c]{6.5}{*}{\rotatebox{90}{VGG16-BN}} & PFEC$^*$ & 91.85 & 91.29 & 0.56 & 13.89 & \xmark \\
				& SSL$^*$ & 92.09 & 91.80 & 0.29 & 17.76 & \xmark\\
				& DMCP & 92.21 & 92.04 & 0.17 & 25.05 & \xmark \\ 
				& FTWT & 93.82 & 93.73 & 0.09 & 44.00 & \xmark \\
				\cmidrule{2-7}
				& \textbf{Ours} ($L=18$) & 94.00 & 92.03 $(\pm 0.21)$ & 1.97 & 37.15 $(\pm 0.47)$ & \cmark \\
				& \textbf{Ours} ($L=22$) & 94.00 & 93.00 $(\pm 0.12)$ & 1.00 & 33.25 $(\pm 0.44)$ & \cmark \\
				\midrule
				\multirow[c]{5.5}{*}{\rotatebox{90}{VGG19-BN}} & PFEC$^*$ & 92.11 & 91.78 & 0.33 & 16.55 & \xmark \\
				& SSL$^*$ & 92.02 & 91.60 & 0.42 & 30.68 & \xmark \\
				& DMCP  & 92.19 & 91.94 & 0.25 & 34.14 & \xmark \\ 
				\cmidrule{2-7}
				& \textbf{Ours} ($L=18$) & 93.95 & 92.32 $(\pm 0.35)$ & 1.63 & 38.83 $(\pm 0.36)$ & \cmark \\
				& \textbf{Ours} ($L=22$) & 93.95 & 93.22 $(\pm 0.14)$ & 0.73 & 34.11 $(\pm 0.99)$ & \cmark \\
				\bottomrule 
			\end{tabular}
			\begin{tablenotes}\footnotesize
				\item[*] Results taken from \citet{Xu2021Efficient}.
			\end{tablenotes}
		\end{threeparttable}%
	}
\end{table}

\begin{table}
	\centering
	\caption{Latency estimates for the \methodname\ module on CIFAR-10. The realistic setting assumes hardware support for efficient patch-wise operations. The theoretical speedup is derived from the achieved FLOPs reduction.}
	\label{tab_latency_c10}
	\resizebox{0.67\linewidth}{!}{%
		\begin{threeparttable}
			\begin{tabular}{cccc}
				\toprule
				Model & Setting & Latency & Speedup \\
				\midrule
				\multirow[c]{3.5}{*}{\makecell{ResNet18 \\ ($L=14$)}} & Baseline & $8.73\,$ms & / \\
				\cmidrule{2-4}
				& Realistic & $5.88\,$ms & $1.48$x \\
				& Theoretical & $5.09\,$ms & $1.72$x \\
				\midrule
				\multirow[c]{3.5}{*}{\makecell{ResNet34 \\ ($L=14$)}} & Baseline & $15.54\,$ms & / \\
				\cmidrule{2-4}
				& Realistic & $10.60\,$ ms & $1.47$x \\
				& Theoretical & $8.28\,$ ms & $1.88$x \\
				\bottomrule 
			\end{tabular}
		\end{threeparttable}%
	}
\end{table}

\subsection{Results on ImageNet}
\label{sec_results_on_imagenet}

\begin{table}
	\centering
	\caption{Selected results on ImageNet. "FLOPs Red." denotes the percentage reduction of FLOPs after pruning compared to the baseline.}
	\label{tab_results_imagenet}
	\resizebox{1\linewidth}{!}{%
		\begin{threeparttable}
			\begin{tabular}{ccccccc}
				\toprule
				\multirow[c]{2.5}{*}{Model} & \multirow[c]{2.5}{*}{Method} & \multicolumn{3}{c}{Top-1 Accuracy (\%)} & \multirow[c]{2.5}{*}{\makecell{FLOPs Red.\ \\(\%)}} & \multirow[c]{2.5}{*}{\makecell{Data-\\Free}}\\
				\cmidrule(lr){3-5}
				& & Baseline & Pruned & $\Delta$ & & \\
				\midrule
				\multirow[c]{7.5}{*}{\rotatebox{90}{ResNet18}} & LCCN & 69.98 & 66.33 & 3.65 & 34.60 & \xmark \\
				& DynConv$^*$ & 69.76 & 66.97 & 2.79 & 41.50 & \xmark  \\ 
				& FPGM & 70.28 & 68.34 & 1.94 & 41.80 & \xmark  \\
				& FBS & 70.71 & 68.17 & 2.54 & 49.49 & \xmark \\
				& FTWT & 69.76 & 67.49 & 2.27 & 51.56 & \xmark \\
				\cmidrule{2-7}
				& \textbf{Ours} ($L=16$) & 69.76 & 66.97 $(\pm 0.21)$ & 2.79 & 18.28 $(\pm 0.19)$ & \cmark \\
				& \textbf{Ours} ($L=20$) & 69.76 & 68.64 $(\pm 0.56)$ & 1.12 & 15.10 $(\pm 0.18)$ & \cmark \\
				\midrule
				\multirow[c]{7.5}{*}{\rotatebox{90}{ResNet34}} & PFEC & 73.23 & 72.09 & 1.14 & 24.20 & \xmark  \\
				& LCCN & 73.42 & 72.99 & 0.43 & 24.80 & \xmark \\
				& FPGM & 73.92 & 72.54 & 1.38 & 41.10 & \xmark \\
				& FTWT & 73.30 & 72.17 & 1.13 & 47.42 & \xmark \\
				& DGNet & 73.31 & 71.95 & 1.36 & 67.20 & \xmark \\
				\cmidrule{2-7}
				& \textbf{Ours} ($L=16$) & 73.31 & 70.31 $(\pm 0.07)$ & 3.00 & 22.65 $(\pm 0.45)$ & \cmark \\
				& \textbf{Ours} ($L=20$) & 73.31 & 72.06 $(\pm 0.05)$ & 1.25 & 18.69 $(\pm 0.30)$ & \cmark \\
				\midrule
				\multirow[c]{4.5}{*}{\rotatebox{90}{ResNet50}} & FPGM & 76.15 & 74.83 & 1.32 & 53.50 & \xmark \\
				& DGNet & 76.13 & 75.12 & 1.01 & 67.90 & \xmark \\
				\cmidrule{2-7}
				& \textbf{Ours} ($L=28$) & 76.13 & 73.04 $(\pm 0.07)$ & 3.09 & 18.58 $(\pm 0.33)$ & \cmark \\
				& \textbf{Ours} ($L=36$) & 76.13 & 74.77 $(\pm 0.10)$ & 1.36 & 15.68 $(\pm 0.16)$ & \cmark \\
				\bottomrule 
			\end{tabular}
			\begin{tablenotes}\footnotesize
				\item[*] Results taken from \cite{Li2021Dynamic}.
			\end{tablenotes}
		\end{threeparttable}%
	}
\end{table}

\begin{table}
	\centering
	\caption{Latency estimates for the \methodname\ module on ImageNet. The realistic setting assumes hardware support for efficient patch-wise operations. The theoretical speedup is derived from the achieved FLOPs reduction.}
	\label{tab_latency_imagenet}
	\resizebox{0.67\linewidth}{!}{%
		\begin{threeparttable}
			\begin{tabular}{cccc}
				\toprule
				Model & Setting & Latency & Speedup \\
				\midrule
				\multirow[c]{3.5}{*}{\makecell{ResNet34 \\ ($L=16$)}} & Baseline & $103.50\,$ms & / \\
				\cmidrule{2-4}
				& Realistic & $84.56\,$ms & $1.22$x \\
				& Theoretical & $80.06\,$ms & $1.29$x \\
				\midrule
				\multirow[c]{3.5}{*}{\makecell{VGG19-BN \\ ($L=20$)}} & Baseline & $476.96\,$ms & / \\
				\cmidrule{2-4}
				& Realistic & $371.59\,$ ms & $1.28$x \\
				& Theoretical & $329.91\,$ ms & $1.45$x \\
				\bottomrule 
			\end{tabular}
		\end{threeparttable}%
	}
\end{table}

\begin{figure}
	\centering
	\begin{subfigure}{0.45\textwidth}
		\centering
		\includegraphics[width=0.95\textwidth]{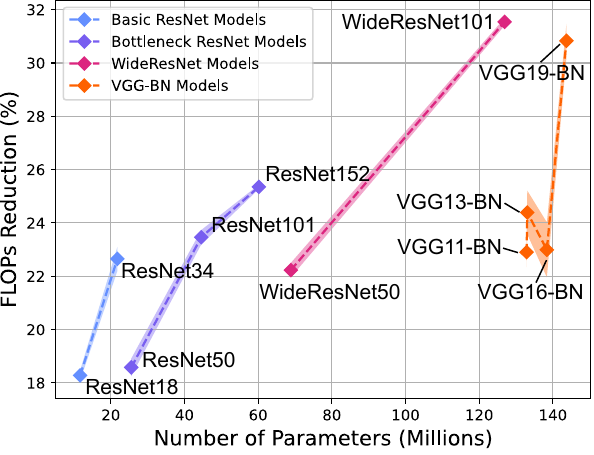}
		\caption{Overview of ImageNet experiments. }
		\label{subfig_imagenet_overview}
	\end{subfigure}
	\par\bigskip\bigskip
	\begin{subfigure}{0.45\textwidth}
		\centering
		\includegraphics[width=0.95\textwidth]{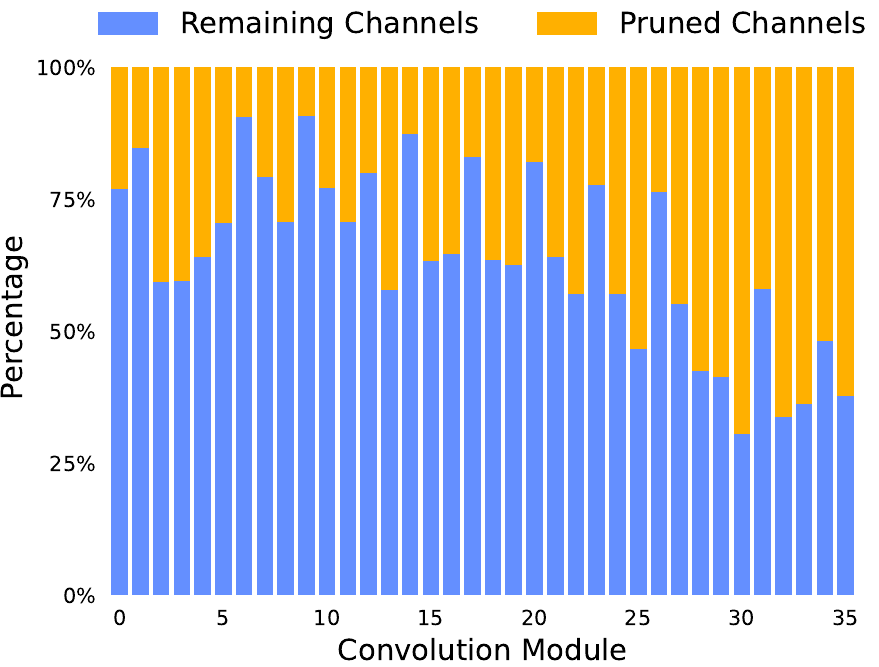}
		\caption{Distribution of pruned channels in a ResNet50.}
		\label{subfig_resnet50_gating}
		\vspace*{0.25cm}
	\end{subfigure}
	\caption{Visualization of results on the ImageNet dataset. (a) depicts the relation of FLOPs reduction to number of parameters for all tested architectures. Results are shown with $L=16$ for basic ResNet models, $L=28$ for bottleneck ResNets, $L=32$ for WideResNets, and $L=20$ for VGG-BN models. (b) shows the achieved compression rate per convolution module in a ResNet50, starting from the second bottleneck layer.}
\end{figure}

On the ImageNet benchmark dataset, we evaluate all available ResNet architectures including WideResNets as well as all VGG-BN models. Results are presented in Figures \ref{subfig_imagenet_overview} and \ref{subfig_resnet50_gating}. In particular, we observe a positive scaling behavior of our method in Figure \ref{subfig_imagenet_overview}, achieving up to 31.54\% FLOPs reduction for a WideResNet101. When observing models of similar architecture, the potential FLOPs reduction grows with the number of parameters. We relate this to the fact that larger models typically exhibit more redundancies, which are then compressed by our module. 

Similar to \citet{He2018AMC}, we observe that models including pointwise convolutions are harder to prune than their counterparts which rely solely on larger filter kernels. This is particularly apparent in the drop in FLOPs reduction from ResNet34 to ResNet50. While the larger ResNet and WideResNet models with bottleneck blocks continue the scaling pattern, the introduction of pointwise convolutions momentarily dampens the computational cost reduction. Increasing the width of each convolutional layer benefits pruning performance, as is apparent with the results of WideResNet50 with twice the number of channels per layer as in ResNet50. While pointwise convolutions can achieve similar or even better compression ratios compared to $3\times 3$ convolutions (see Figure \ref{subfig_resnet50_gating}), the cost overhead of the hashing and merging steps is higher relative to the baseline.

When comparing the results to those on CIFAR-10, we note that our \methodname\ module achieves less compression on ImageNet classifiers. We directly relate this to the higher complexity in the data. With a 100-fold increase in number of classes and roughly 26 times more training images than on CIFAR-10, the models store more information in latent feature maps, rendering them less redundant and therefore harder to compress. Methods that exploit training data for extensively tuning their pruning scheme naturally achieve higher degrees of FLOPs reduction, as shown in Table \ref{tab_results_imagenet}. However, this is only possible when access to the data is granted. In contrast, our method offers significant reductions of computational cost while being data-free, even scaling with larger model architectures.

%% file: 5_conclusion.tex
\section{\uppercase{Conclusion}}
While existing channel pruning approaches rely on training data to achieve notable reductions in computational cost, our proposed \methodname\ module removes restrictive requirements on data availability and compresses CNNs without requiring any training steps. By employing a locality-sensitive hashing scheme for redundancy detection, we are able to drastically reduce the depth of latent feature maps and corresponding convolutional filters to significantly decrease the model's total FLOPs requirement. Our approach prunes the model at runtime in an input-dependent manner, even allowing for changes to the compression ratio in real time. 
This property could be particularly suitable in a federated learning scenario, where the model's weights are continuously updated, rendering other pruning methods which require pre-processing of the model's weights infeasible. 

We empirically validate our claim through a series of experiments with a variety of CNN models and achieve compelling results on the CIFAR-10 and ImageNet benchmark datasets. We aim for our method to serve as an initial step in the direction of entirely data-free  methods for on-the-fly compression of convolutional architectures. 
Future work involves the integration of our method into related computer vision tasks and its extension to novel architectures.

%% file: 6_appendix.tex
\section*{Latency Considerations}
\label{appendix_latency}
While many pruning approaches focus on generating small but dense models that are easy to execute, it is also possible to achieve significant latency benefits using methods that leverage non-contiguous sets of weights which are chosen in an input-dependent manner \citep{Chen2020SLIDE, Chen2021MONGOOSE,Kitaev2020Reformer, Belcak2023Exponentially}. Our \methodname\ module employs a similar technique by only computing the convolution on non-redundant channels. 

However, modern deep learning frameworks do not support conditional execution operations natively \citep{Belcak2023Exponentially} and are optimized towards large, dense matrix multiplications, as is the case with PyTorch \citep{Paszke2019PyTorch}.
Thus, highly optimized implementations are necessary to allow conditional execution strategies to compete with dense models. We focus our efforts on providing a proof of concept for the viability of dynamic, data-free pruning in PyTorch due to its wide-spread use in machine learning research. 

For the latency estimates shown in Tables \ref{tab_latency_c10} and \ref{tab_latency_imagenet} of the main text, we present two different scenarios:
\begin{itemize}
	\item \textbf{Realistic.} In this scenario, we assume that the hardware is capable of handling patch-wise varying channel depths. This allows for accurate execution of our proposed method, as different compression ratios per patch can be fully utilized.
	\item \textbf{Theoretical.} In the theoretical setting, we assume that the latency of the baseline model is reduced by the same amount as the reduction in FLOPs, as observed in our experiments. 
\end{itemize}
In both scenarios, we measure the total latency per image of the model equipped with our proposed \methodname\ modules, across a batch of 64 images from the respective dataset. Since the PyTorch framework does not support efficient computations with ternary weights $\{-1, 0, 1\}$ as required for our hashing scheme, we extrapolate its latency based on the FLOPs count.

\section*{Pruning Pointwise Convolutions}
\label{appendix_1x1_conv}
A special case of the convolution operation appears when $K=1$. These $1\times1$ convolutions are commonly used for downsampling or upsampling of the channel dimension before and after parameter-heavy convolutions with larger kernel sizes, or after a depth-wise convolutional layer.
However, as the kernel resolution changes to a single pixel, each input pixel generates exactly one output pixel in the spatial domain. As there is no reduction in spatial resolution when performing $1 \times 1$ convolutions, we do not require the $3 \times 3$ patches that rasterize the input to be overlapping. Hence, we pad the input in such a way that each side is divisible by $3$ and use non-overlapping patches.

\section*{Component Ablation}
\label{appendix_ablation}

To put the results of our LSH-based data-free compression method into context, we construct an ablation study which analyzes the impact of our method's individual components. As a baseline for comparison, we employ an $L^1$ norm-based pruning criterion and apply it in various settings to establish a fair comparison to our proposed \methodname\ module. For all experiments we compute the $L^1$ norm of channels of the input feature maps of convolution modules and prune a fixed percentage of channels with the lowest norm (see \citep{Li2017Pruning}) to achieve comparable FLOPs reductions to the \methodname\ module.

The results are presented in Tables \ref{tab_ablation_overview} and \ref{tab_ablation_results}. At a given compression ratio, the $L^1$ norm-based pruning approaches do not keep the pruned model's accuracy at an acceptable level. In contrast, the proposed HASTE module is able to keep near-baseline accuracy.

\begin{table}
	\centering
	\caption{Overview of experiments for the data-free $L^1$ norm-based pruning baseline. "Usage of Patches" denotes whether the pruning is applied to an entire input channel (\xmark), or individually for each channel of each patch (\cmark), as visualized in Figure \ref{figure_method} of the main text. "Pruning Criterion" indicates whether the $L^1$ norm of channels or locality-sensitive hashing (LSH) is used to determine which channels to prune. Lastly, "Pruning Operation" denotes if the selected channels are removed or merged into one singular channel.}
	\label{tab_ablation_overview}
	\resizebox{1\columnwidth}{!}{%
		\begin{tabular}{cccccc}
			\toprule
			& Prune & Merge & Patch-Prune & Patch-Merge & Ours \\
			& (P) & (M) & (PP) & (PM) & (\methodname) \\
			\midrule
			Usage of Patches & \xmark & \xmark & \cmark & \cmark & \cmark \\
			\midrule
			Pruning Criterion & $L^1$ & $L^1$ & $L^1$ & $L^1$ & LSH \\
			\midrule
			Pruning Operation & Remove & Merge & Remove & Merge & Merge \\
			\bottomrule 
		\end{tabular}%
	}
\end{table}

\begin{table}
	\centering
	\caption{Comparison of results of data-free $L^1$ norm-based pruning methods (see Table \ref{tab_ablation_overview}) to our proposed \methodname\ module on the CIFAR-10 dataset. ”FLOPs Red.” denotes the percentage decrease of FLOPs after pruning compared to the base model. We highlight the highest remaining Top-1 accuracy and lowest loss of accuracy ($\Delta$) for each compression target in \textbf{bold}.}
	\label{tab_ablation_results}
	\resizebox{1\columnwidth}{!}{%
		\begin{tabular}{cccccc}
			\toprule
			\multirow[c]{2.5}{*}{Model} & 
			\multirow[c]{2.5}{*}{Method} &
			\multicolumn{3}{c}{Top-1 Accuracy (\%)} & 
			\multirow[c]{2.5}{*}{\makecell{FLOPs \\ Reduction (\%)}} \\
			\cmidrule{3-5}
			& & Baseline & Pruned & $\Delta$ & \\
			\midrule
			\multirow[c]{5}{*}{ResNet18} & P & 93.07 & 71.07 & 22.00 & 40.80 \\
			& M & 93.07 & 65.31 & 27.76 & 41.75 \\
			& PP & 93.07 & 88.70 & 4.37 & 40.80 \\
			& PM & 93.07 & 86.53 & 6.54 & 39.89 \\
			& \methodname\ & 93.07 & \textbf{91.18} & \textbf{1.89} & 41.75 \\
			\midrule
			\multirow[c]{5}{*}{ResNet34} & P & 93.34 & 48.42 & 44.92 & 51.98 \\
			& M & 93.34 & 40.52 & 52.82 & 53.13 \\
			& PP & 93.34 & 80.04 & 13.30 & 51.98 \\
			& PM & 93.34 & 72.10 & 21.24 & 50.51 \\
			& \methodname\ & 93.34 & \textbf{90.45} & \textbf{2.89} & 51.09 \\
			\midrule
			\multirow[c]{5}{*}{VGG11-BN} & P & 92.39 & 41.77 & 50.62 & 37.87 \\
			& M & 92.39 & 73.87 & 18.52 & 38.90 \\
			& PP & 92.39 & 65.94 & 25.45 & 37.87 \\
			& PM & 92.39 & 87.39 & 5.00 & 37.11 \\
			& \methodname\ & 92.39 & \textbf{89.36} & \textbf{3.03} & 37.25 \\
			\midrule
			\multirow[c]{5}{*}{VGG19-BN} & P & 93.95 & 34.89 & 59.06 & 40.73 \\
			& M & 93.95 & 42.23 & 51.72 & 42.02 \\
			& PP & 93.95 & 65.84 & 28.11 & 40.72 \\
			& PM & 93.95 & 82.51 & 11.44 & 40.31 \\
			& \methodname\ & 93.95 & \textbf{91.19} & \textbf{2.76} & 41.47 \\
			\bottomrule 
		\end{tabular}%
	}
\end{table}

\section*{Visualizations}
\label{appendix_visualizations}
\begin{figure*}
	\centering
	\includegraphics[width=1.0\linewidth]{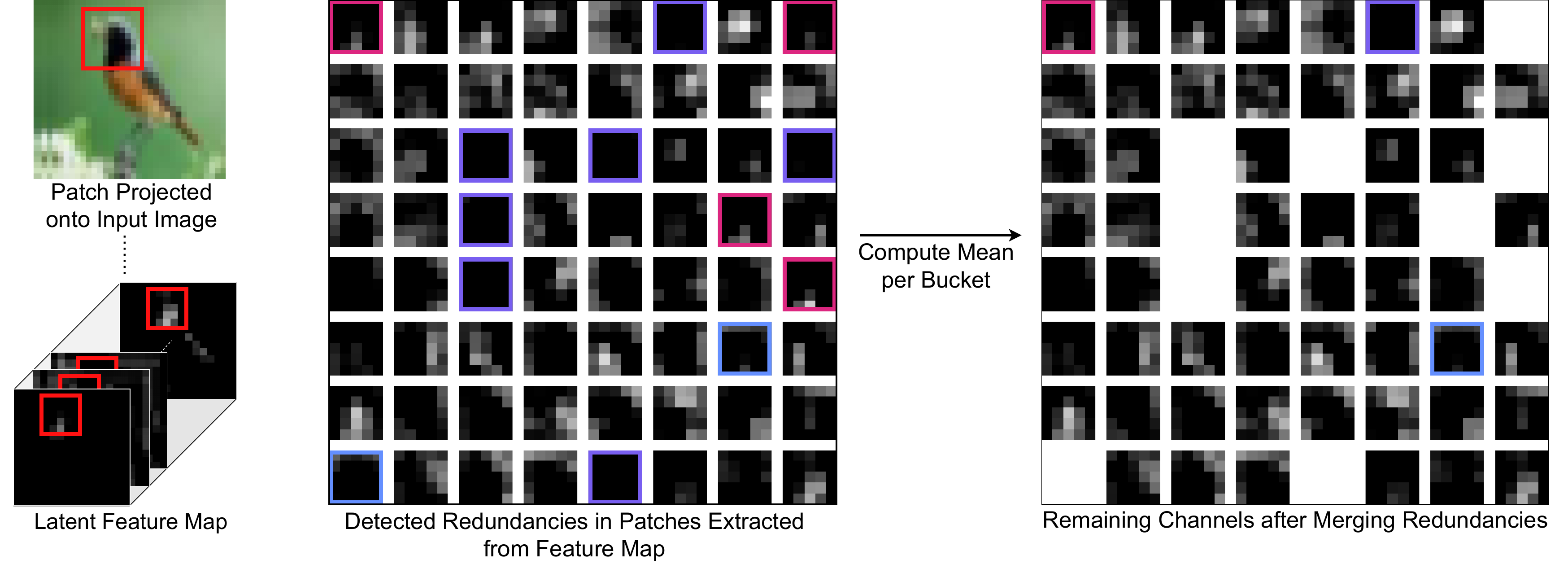}
	\caption{Visualization of the input channel compression performed by the \methodname\ module. Patches with identical hash codes receive identical outline colors and are averaged by taking their mean. Patches with no matching hash code are left unchanged. Here, we reduce $C_{in} = 64$ to $\tilde{C}_{in} = 54$, which gives us a compression ratio of $r = 15.63\%$.}
	\label{fig_fm_vis_2}
\end{figure*}

\begin{figure*}
	\centering
	\includegraphics[width=1.0\linewidth]{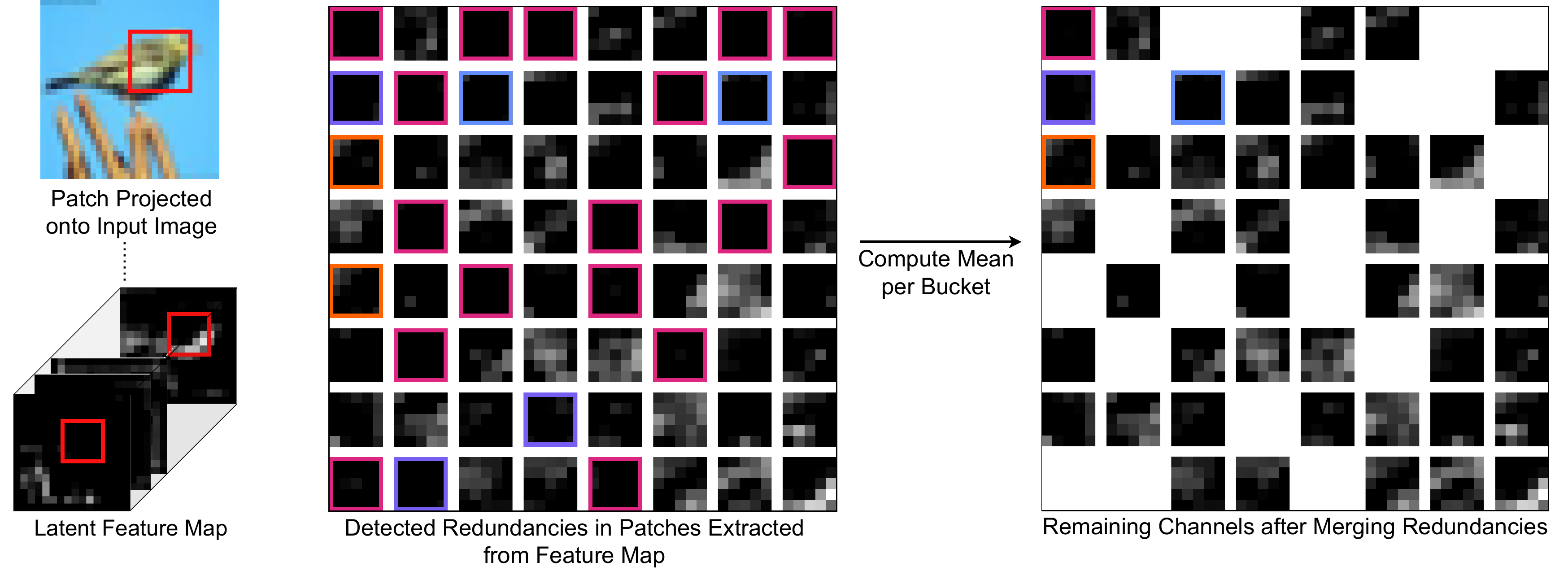}
	\caption{Visualization of the input channel compression performed by the \methodname\ module. Patches with identical hash codes receive identical outline colors and are averaged by taking their mean. Patches with no matching hash code are left unchanged. Here, we reduce $C_{in} = 64$ to $\tilde{C}_{in} = 44$, which gives us a compression ratio of $r = 31.25\%$.}
	\label{fig_fm_vis_3}
\end{figure*}

\begin{figure*}
	\centering
	\includegraphics[width=1.0\linewidth]{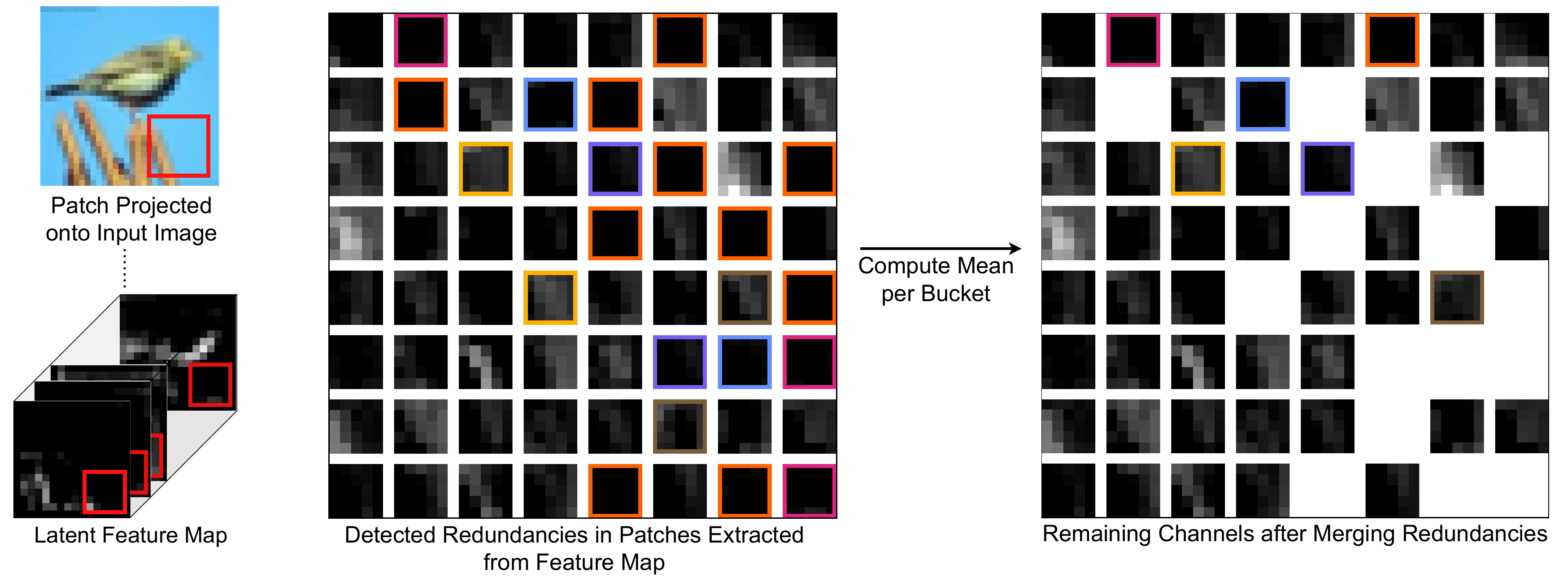}
	\caption{Visualization of the input channel compression performed by the \methodname\ module. Patches with identical hash codes receive identical outline colors and are averaged by taking their mean. Patches with no matching hash code are left unchanged. Here, we reduce $C_{in} = 64$ to $\tilde{C}_{in} = 49$, which gives us a compression ratio of $r = 23.43\%$.}
	\label{fig_fm_vis_4}
\end{figure*}

To gain an intuitive understanding of the merge operation for redundant feature map channels as described in Section \ref{method_our_module} of the main text, we provide visualizations of the latent features before and after the merging step in Figures \ref{fig_fm_vis_1}, \ref{fig_fm_vis_2}, \ref{fig_fm_vis_3} and \ref{fig_fm_vis_4}. 
Note that the compression ratio $r = 1-(\tilde{C}_{in} / C_{in}) \in (0, 1)$ changes not only depending on the input image, but on the amount of redundancies found in each individual patch. The comparison of Figures \ref{fig_fm_vis_1} and \ref{fig_fm_vis_2} reveal an interesting property of our proposed \methodname\ module: Patches that contain little class-specific information, such as the background, can be compressed to a much higher degree than patches that contain relevant information for the classification task. 

